# Unknown Words Analysis in POS tagging of Sinhala Language

A.J.P.M.P. Jayaweera[#1], N.G.J. Dias[*2]

[#] *Virtusa Pvt. Ltd.*
*No 752, Dr. Danister De Silva Mawatha, Colombo 09, Sri Lanka*
[1] `mjayaweera@gmail.com`

[*] *Department of Statistics & Computer Science, University of Kelaniya*
*Kelaniya, Sri Lanka*
[2] `ngjdias@kln.ac.lk`

*Abstract*— Part of Speech (POS) is a very vital topic in Natural Language Processing (NLP) task in any language, which involves analysing the construction of the language, behaviours and the dynamics of the language, the knowledge that could be utilized in computational linguistics analysis and automation applications. In this context, dealing with unknown words (words do not appear in the lexicon referred as unknown words) is also an important task, since growing NLP systems are used in more and more new applications. One aid of predicting lexical categories of unknown words is the use of syntactical knowledge of the language. The distinction between open class words and closed class words together with syntactical features of the language used in this research to predict lexical categories of unknown words in the tagging process. An experiment is performed to investigate the ability of the approach to parse unknown words using syntactical knowledge without human intervention. This experiment shows that the performance of the tagging process is enhanced when word class distinction is used together with syntactic rules to parse sentences containing unknown words in Sinhala language.

## I. INTRODUCTION

Part of speech tagging is one of the pivotal steps in the knowledge acquiring process in natural language processing task. The fundamental processing step in tagging consists of assigning POS tags to every token in the text with a corresponding POS tag like noun, verb, preposition, etc., based both on its definition, as well as its context. Appearance of an unknown is the one of the problems that is facing in natural language parsing systems, i.e., the words that appear in sentences, but are not contained within the lexicon. New words are continually coined to the language, and people will often use words is parsing, that the system may not expect. This problem get worse when NLP systems are used for more and more on-line computer applications.

This paper will discuss how well a distinction of Sinhala word classes, syntactic rules can be used in parsing sentences containing unknown words in natural language processing tasks. The distinction between closed class and open class words should help to refine the possibilities for unknown words, then syntactic knowledge can be used to aid in the analysis of unknown words sentence structure, which can be a strong evidence for the possible part of speech of an unknown word. We expect that these two knowledge sources will greatly improve tagging ability to process and handle with words that are not in the system corpus.

In this paper, we presents importance of handling unknown word in part of speech tagging process, and an approach is suggested. Section II of this paper gives an idea of the background of the problem and details of previous research. Section III describes distinction of open class and closed class words in Sinhala language, and section IV gives details of morphological and syntactical analysis of the current text corpus. Section V discusses about the approach that we have proposed for guessing parts of speech for unknown words. Section VI and VII discuss the Evaluation, testing and the results, and section VIII concludes the paper and describes the future work.

## II. BACKGROUND OF THE PROBLEM OF UNKNOWN WORDS

Appearance of unknown words is one of the frequently occurring problems facing in part of speech tagging process, i.e., the words that appear in sentences, but are not contained within the training corpus. New words are continually entering the language, Acronyms and proper names are created very often and new nouns and verbs are adding to the language in a surprising rate. So it is impossible to train the tagger for every possible word in the language. So unknown words are non-negligible in POS tagging. Therefore, in order to build a complete tagger, tagger must be incurred with some knowledge of suggesting the tag for an unknown word.

There are two approaches to handle unknown words. The first approach is to attempt to construct a complete lexicon, then deal with unknown words in a simple way. For example, rejecting the input. The second approach is to attempt to analyse the word at the time of encounter with using a set of human defined rules. This would allow the tagger to process sentences containing unknown words.

Before examining the problem in detail, it is useful to consider work that has already been done by other researchers. There have been several attempts to study the problem of learning unknown words. These attempts have followed several different methodologies and have focused on various aspects of the unknown words.

Previous techniques reported for other languages such as English, have mostly utilize the guessing rules to analyse the word features by looking at leading and trailing characters. Most of them employ the analysis of trailing characters and other features such as capitalization and hyphenation. Some of them use more morphologically oriented word features such as suffixes, prefixes, and character lengths. The guessing rules are usually use knowledge of morphology of the language.

The simple possible way that suggested [2] is to consider each unknown word that is ambiguous among all possible tags, with equal probability, and then using contextual POS-trigram from the corpus to suggest the proper tag.

There are more complex methods, which have been tried out by other researchers for dealing with unknown words using morphological and syntactical features of the language. Eric Brill [3] make use of morphology to handle unknown words during part of speech tagging process. Brill's tagger begins by tagging unknown words as proper noun if capitalized, as common noun if not. Then the tagger learns various transformational rules in the training process from the tagged corpus. Then it applies these rules to unknown words, to tag with the appropriate parts of speech category. Scott M. Thede and Mary Harper [5] in their paper presented an approach using morphology and syntactic parsing rules in post-mortem method for determining the probable lexical classes of words. Tetsuji, Taku Kudoh and Yuji [6] proposed a POS tagging approach for unknown English words using Support Vector Machines (SVM). SVM classifiers are created for each POS tag using all words in the training set, then POS tags to unknown words predict using those classifiers.

But an agglutinative language presents more serious problems with unknown words, unlike English. Gary, Jeongwon, Jong-Hyeok [4] have proposed a syllable-pattern-based generalized unknown-morpheme estimation method using a morpheme pattern dictionary in their statistical and rule-based hybrid POS tagging system for Korean language.

Since Sinhala is also a complex, morphologically rich and agglutinative language, information about morphology or how word is spelled is very difficult to use in unknown word prediction algorithms.

### III. OPEN CLASS VS CLOSED CLASS WORDS

Traditionally, the definition of POS is based on morphological and syntactic functions. Similar to most of other languages, POS in Sinhala language also can be divided into two broad categories: closed class type and open class type. Closed classes are those that have relatively fixed membership. Closed class words are generally function words: which tend to be very short, occur frequently, and play an important role in grammar. By contrast, open class is the type that lager number of words are belong in any language, and new words are continually coined or borrowed from other languages. The words that are usually containing main content of a sentence are belonged to open word class category.

In Sinhala, all Nouns and Verbs can be categorized under open word class. But Nipatha and Upasarga behave differently in Sinhala grammar. Words belong to Nipatha and Upasarga are not changed according to time and gender, Upasarga always join with nouns and provide additional (improved) meaning to the noun, therefore, Upasarga are not categorized under any of word class, but Nipatha can be categorized as closed class words based on their existence.

In-addition to that, Sinhala Pronouns can be classified as open class words, based on their morphological properties, but Pronouns also can be classified as closed class words, based on their existence of fixed membership in the language.

### IV. ANALYSIS OF MORPHOLOGICAL AND SYNTACTICAL FEATURES

NLP based language analysis mainly aid by morphological and syntactical features analysis of a language, and the availability of lexical resources is essential in this tasks. So having a corpus for a language is an important lexical resource in the field of NLP. In order to make the corpora more useful for doing linguistic research, they are required to annotate with respective knowledge sources to make it suitable to process with linguistics applications. One example of annotating a corpus is part of speech tagging, in which information about each word's part of speech (verb, noun, adjective, etc.) is added to the corpus in the form of tags.

The corpus that we use in this research is the beta version of the Corpus developed by the UCSC under PAN Localization project in 2005 [7], which contains around 2754 sentences and 90551 words tagged with corresponding part of speech tag, that comprise of data drawn from different kinds of Sinhala newspaper articles under different classifications, mainly form Art, Sports, Science, Indigenous knowledge and Religion.

We performed a detailed analysis of the corpus to understand what rules govern the language and what patterns occur. Empirical results of the analysis is described in this paper in detail. A substantial effort was made at the corpus preparation phase to correct issues encountered in the formatting of the text in the corpus.

In order to further analysis of the corpus, word frequency distribution and the tag frequency distribution were obtained from the corpus by running a simple tokenizing program. Tags with typographical errors and irrelevant tokens (numbers, foreign words, etc.) were removed from the list after manual inspection.

*A. Most Frequent Words in Sinhala Text Corpus*

This analysis was performed to observe the words and Part of Speech categories that are more frequent in Sinhala text corpus. In order to observe most frequent words, a distinct word list with frequencies were obtained from the corpus along with possible part of speech tags. For simplicity, only top 20 words were considered in the analysis. Table I contains the list of words, with frequencies and possible part of speech categories.

It is observed that, most frequent words in Sinhala language are function words, which belong to closed class category. 11 out of 20 words belong to Nipatha which all are function words, 6 frequent verbs are also within the list. It is also observed that tagging ambiguity exists among high frequent words, though they are function words. There are two words among top 20 words, "පත්" and "යුතු" that are not properly classified into respective parts of speech categories.

*B. Zipf's Law Analysis*

Zipf's Law states that, the frequency of occurrence of an instance of a class is roughly inversely proportional to the rank of that class in the frequency list, for example occurrences of words in a document. So the goal of this test was to observe parts of speech distribution within Sinhala language displays the Zipf's Law behavior.

Suppose that, a word occurs $f$ times and that in the list of word frequencies it has a certain rank $r$, if Zipf's Law holds we have (for all words) $f = a/r^b$ where $a$ and $b$ are constants and $b$ is close to minus 1 (-1). Taking the logarithm of each side of the equation we get:

TABLE I

THE TOP 20 MOST FREQUENT WORDS IN SINHALA LANGUAGE (REFER APPENDIXES 1 FOR DESCRIPTION OF EACH TAG)

| Word | Frequ. | Possible POS tags | Main POS Category/s |
|---|---|---|---|
| අ | 1089 | POST (1024), VNF (57), VNN, VP, RP | Nipatha - Postposition |
| ඇති | 632 | VP (621), JVB (1) | Nipatha - Verb Participle |
| අතර | 567 | POST (376), CC (191) | Nipatha - Postposition |
| හා | 530 | CC (526), POST (3) | Nipatha - Conjunctions |
| වූ | 436 | VP (430), VNF (6) | Nipatha - Verb Participle |
| ඇත | 429 | VFM (409), VP (19), VNF | Verb Finite Main |
| සහ | 401 | CC (401) | Nipatha - Conjunctions |
| මෙම | 370 | DET (365), POST (5) | Nipatha - Determiner |
| වී | 355 | VNF (340), VP (6), NNN, VNN, NNPA | Verb Non Finite |
| ඔහු | 348 | PRP (347), POST (1) | Pronoun |
| කළ | 331 | VP (304), VNF (24), VFM, NNN | Nipatha - Verb Participle |
| විය | 313 | VFM (194), VP (111), VNF, RP, NNN | Verb Finite Main, Verb Participle |
| කර | 307 | VNF (302), VP (2), VFM, VNN | Verb Non Finite |
| පත් | 290 | ? (259 times), VNF (13), NNN (11), VP, JVB | Not properly classified |
| ගෙන | 283 | VNF (244), VP (37), POST | Verb Non Finite |
| කැර | 275 | VNF (271), VP (3), VNN | Verb Non Finite |
| විසින් | 274 | POST (264) | Nipatha - Postposition |
| තුළ | 224 | POST (222) | Nipatha - Postposition |
| යුතු | 217 | ? (191 times), POST (25) | Not properly classified |
| එම | 213 | DET (207), PRP (5), NNPA | Nipatha - Determiner |

$$\log(f) = \log(a) - b * \log(r)$$

Table II contains tag frequency occurred in the corpus and ordered in descending order, rank one was assigned to the top most one. The Fig. I shows the plot of $\log(r)$ vs $\log(f)$ calculated for the data presented in the Table II. From Fig. I we can observe that the tag frequencies are roughly form a line from the upper-left corner to the lower-right corner of the graph with slope close to -1. This indicates that the parts of speech distribution of Sinhala language is also displays Zipf's Law behavior.

TABLE II

THE TAG FREQUENCY WITH RANKING

| Tag | Frequency (f) | Rank (r) |
|---|---|---|
| NNN | 21034 | 1 |
| VP | 7257 | 2 |
| NNPI | 6651 | 3 |
| POST | 5385 | 4 |
| JJ | 5215 | 5 |
| VNF | 4735 | 6 |
| RP | 4181 | 7 |
| NNM | 3895 | 8 |
| NNPA | 3604 | 9 |
| NVB | 3303 | 10 |
| VNN | 2560 | 11 |
| VFM | 2475 | 12 |
| PRP | 2421 | 13 |
| DET | 2014 | 14 |
| QFNUM | 1820 | 15 |
| CC | 1763 | 16 |
| JVB | 785 | 17 |
| RB | 699 | 18 |
| NNF | 368 | 19 |
| FRW | 154 | 20 |

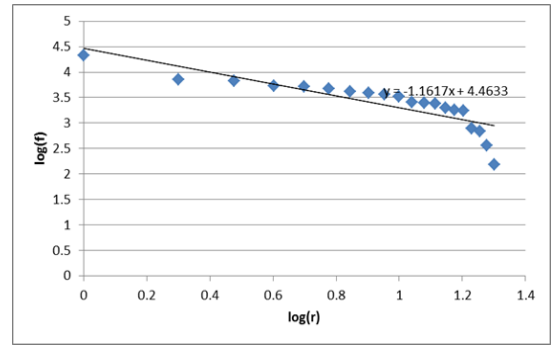

Fig. I Plot of log(r) versus log(f)

*C. Maximum Likelihood Estimate of the Tag Distribution*

The objective of this test is to observe the Likelihood distribution of parts of speech tags in Sinhala text corpus and understand which tags are most likely to appear. Table III contains tag distribution obtained from the corpus. Tag frequency and number of distinct words with respect to each tag were counted from the corpus. The Likelihood Estimates were calculated for each tag based on number of occurrences of the tag appeared in the corpus against total number of tags in the corpus.

Maximum Likelihood Estimation (MLE) was calculated using,

$$Maximum\ Likelihood\ Estimate = \frac{c(t)}{c(w)},$$

TABLE III
TAG FREQUENCY IN THE CORPUS

| Tag | Number of distinct words | Frequency of the Tag |
|---|---|---|
| **VP** | 1094 | 7257 |
| **RB** | 148 | 699 |
| **RP** | 160 | 4181 |
| **DET** | 107 | 2014 |
| **PRP** | 211 | 2421 |
| **NNF** | 183 | 368 |
| **JVB** | 383 | 785 |
| **NNPA** | 1429 | 3604 |
| **VFM** | 508 | 2475 |
| **NNN** | 6438 | 21034 |
| **FRW** | 111 | 154 |
| **NNM** | 1387 | 3895 |
| **NNPI** | 1723 | 6651 |
| **NVB** | 1057 | 3303 |
| **JJ** | 1324 | 5215 |
| **VNF** | 774 | 4735 |
| **VNN** | 830 | 2560 |
| **QFNUM** | 738 | 1820 |
| **POST** | 310 | 5385 |
| **CC** | 42 | 1763 |

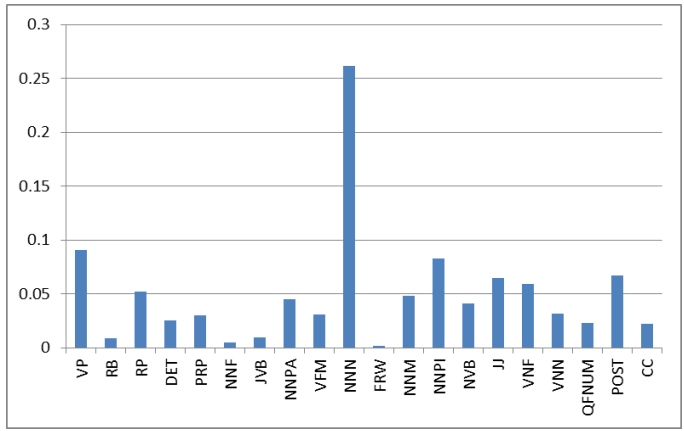

Fig. II Likelihood estimation of Tag distribution

where c(t) is the count of a particular tag and c(w) is the total number of words in the corpus.

According to Fig. II, NNN (Common Noun Neuter) seems to be the most frequent tag in the corpus and it seems to appear about three times more frequently than VP (Verb Participle). It is also noticed that more than 50% of words are noun in the corpus, around 10% belongs to verbs/verbal parses. So in general, we can say around 60% of words in Sinhala language are belong to open class type while around 40% of words belongs to the closed class type.

*D. Words distribution by Tag*

The objective of this test is to analyze, what type of words are exist mostly in Sinhala language. Fig. III is plotted based on data presented in Table III, which shows the distribution of number of distinct words by POS category in the corpus. It is observed that most of the words in Sinhala language are nouns, which is 13047 words are nouns out of total of 18957 distinct words in the corpus, that is almost around 68% of the total number of words in the corpus.

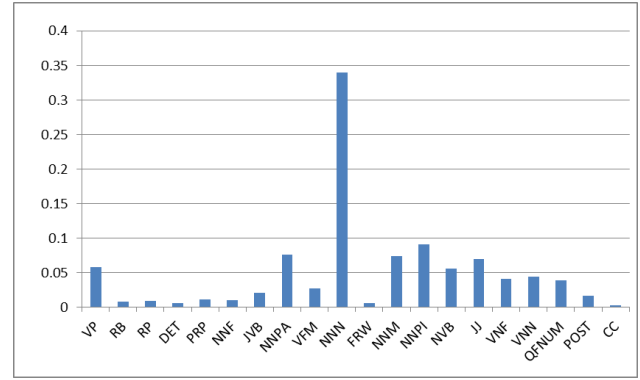

Fig. III Plot of word distribution by Tag

V. GUESSING PART OF SPEECH FOR UNKNOWN WORDS

Tagging data with unknown words is also an essential task in the tagger. When the system reach an unknown word, the initial version of the tagger doesn't cater unknown words, thus tagger fails to propose a tag since the system is not trained for that word and the tagging algorithm doesn't have enough intelligence to propose tags for untrained words.

Since Sinhala is a complex, morphologically rich and agglutinative language, in which information about morphology or how word spelled is very difficult to use in unknown word prediction algorithms, unlike English. So in our research, the important source of information that we have used the distribution of words and parts of speech. So improvements of the algorithm focused on words belong to sub categories of open class words, such as noun, verbs and pronouns. Due to fixed number of membership in closed class word category, we can assume that the words belong to closed class category are well defined in Sinhala grammar and that is fixed. Hence, improvement could be done by incurring knowledge of distinction between closed class and open class words. Syntactic knowledge can be used to aid in the analysis of unknown words sentence structure. Then suggest corresponding part of speech and calculate the trigram using Hidden Markov Model (HMM) for unknown words.

Considering the fact that the tag likelihood distribution and the word frequency distribution of the language, it is clear that NNN is the most frequent part of speech in the language. The NNN tag likelihood is the highest likelihood and it is 3 times greater than VP. By considering word frequency distribution by tags in Fig. III provide evidence that the most of the words in the language are NNNs, which is 5 times more than NNPI. Based on above two factors, the simplest way that could handle unknown words, we can assume and guess NNN as part of speech tag for each new word encountered in the tagging process.

More advanced approach of guessing parts of speech is, the consideration of the distinction between closed class and open class words. This distinction can simplified the unknown prediction algorithm by distinguishing the syntactical categories into separate groups that can greatly simplify the task of processing unknown words in a sentence.

Closed class parts of speech are those that have relatively fixed membership in the language and that may not normally be assigned to new words. Closed class words are words with a closed class part of speech. For example, words belong to Nipatha in Sinhala language are the member of closed class, which has fixed membership in the language, it is very rare that a new Conjunctions (CC), Determiners (DET) or Postpositions (POST) are added to the language. In addition to that words belong to Pronoun (PRP) are also considered as fixed in its existence in the language, and new Pronouns are very rarely created. So, Pronouns are also considered as closed class parts of speech in this research. Table IV contains a list of parts of speech tags that are considered as closed class categories in this research and will be avoided in guessing for unknown words.

Based on morphological and syntactical features of Sinhala language, open class words are comprised of words with the following parts of speech: nouns, verbs, Noun in Kriya Mula, and Adjective in Kriya Mula. Adjectives, adverbs and Verb Participle are also considered under this classification in this research (Table V contains the complete list of parts of speech tags considered as open class category), since those words are syntactically used to modify nouns and verbs.

TABLE IV

LIST OF CLOSED CLASS PARTS OF SPEECH CATEGORIES

| Tag | Part of Speech |
|---|---|
| PRP | Pronoun Common |
| DET | Determiner |
| RP | Particle |
| POST | Postpositions |
| CC | Conjunctions |

In this research, we assume that each new word encountered in the part of speech tagging process of Sinhala language belong to open word class category, hence the unknown word algorithm is to pretend that each unknown word is ambiguous among all open class part of speech tags, with equal probability. Then the tagger computes the tag sequence probability and maximum likelihood probabilities rely on text corpus and suggest the proper tag. Further, foreign words and numerals were handled separately, since they do not form any syntactical relationship with other part of speech categories in the language.

VI. EVALUATION

The evaluation of the system was mainly driven by training the system using the Sinhala text corpus that comprised of 2754 sentences and 90551 words, in which data drown from Sinhala newspaper articles from various genres.

For the test data set, data were carefully selected aiming for testing three different versions of the tagger. The first set was collected from the training corpus aiming to test Version 1 of the tagger that comprised of 621 words only known to the system, and 36 sentences. The second test set was also collected aiming to test Version 2 and 3 of the tagger, which data were drown from Sinhala text corpus outside from the trained data set. The second test set comprised of 51 sentences, 1024 words. Out of 1024 words, 171 words were unknown to the system. To evaluate the performance of the tagger, two gold standard test sets were created.

TABLE V

LIST OF OPEN CLASS PARTS OF SPEECH TAGS

| Tag | Part of Speech |
|---|---|
| NNM | Common Noun Masculine |
| NNF | Common Noun Feminine |
| NNN | Common Noun Neuter |
| NNPA | Proper Noun Animate |
| NNPI | Proper Noun Inanimate |
| VFM | Verb Finite Main |
| VNF | Verb Non Finite |
| VNN | Verbal Non Finite Noun |
| NVB | Noun in Kriya Mula |
| JVB | Adjective in Kriya Mula |
| JJ | Adjective |
| RB | Adverb |
| VP | Verb Participle |

The tagger evaluated by comparing the tagged output with the Gold standard test set. The accuracy was calculated using number of correct tags proposed by the system and total number of words in the sentence/s, by the following formula.

$$\text{Accuracy} = \frac{\text{No. of single correct tags}}{\text{Total no. of words}} * 100\%$$

VII. RESULT AND DISCUSSION

The performance of the tagger was measured, using three different versions of progressively upgraded tagging mechanisms. Version 1 is the simplest form of the tagger that performs well only with known words that rejects all unknown words as tagging failures. Version 2 is a somewhat upgraded version that treats all unknown words as nouns and suggest NNN to each new word encountered in the tagging process. Version 3 is the full version of the tagger that uses statistical technique to guess the best tag for unknown word by considering the context of surrounding words.

Table VI shows our system's performance for these three different versions of tagging mechanisms and the experiment verifies the effectiveness of our unknown word guessing techniques. As shown by the performance of each tagging approach, Version 2 shows a drop in the accuracy, with compared to other two versions. That indicates the approach followed in Version 2 in guessing parts of speech for unknown words is not a reliable method for Sinhala language. Version 3 shows 91.50% of accuracy, that shows considering the distinction between closed and open class in guessing part of speech for unknown words is proved to be useful and an effective way for Sinhala language.

Fig. IV presents the confusion matrix, which summarized the performance of Version 3 of the tagger, where row labels indicate the correct tags and column labels indicate the tags predicted by the system. In this confusion matrix, all correct predictions are located in the diagonal of the table. Most of the deviations are shown in predicting tags for words belong to Common Noun Neuter category.

TABLE VI
PERFORMANCE OF THE TAGGER

| Approach | Tagging Approach | Performance of the Tagger |
|---|---|---|
| Version 1 | Only with known words | 91.30% |
| Version 2 | All unknown words considered as Common Noun Neuter (NNN) | 89.73% |
| Version 3 | Consider distinction between closed class and open class in guessing unknown words | 91.50% |

However, the overall accuracy of the tagger have shown that the distinction between closed class and open class word category is a power full tool in handling unknown words for Sinhala language. But the tagging accuracy is still close to 92%, and that shows more work need to be carried out to fine tune the accuracy of the tagger. So improvement can be suggested to the tagger in handling unknown words. Mainly morphological recognition can also be helpful in predicting possible parts of speech for many unknown words. So the overall performance of the tagger can be improved by using a hybrid approach, with incurring above knowledge to the system by set of hand written rules. Further to make sure this approach more accurate, the output generated by the system need to be manually verified, and retraining the tagger to make sure if the word encountered again later, that word would get properly tagged.

|  | POST | NVB | VP | JJ | RB | RP | VNF | DET | VNN | PRP | NNF | JVB | NNPA | VFM | NNN | NNM | NNPI | CC |
|---|---|---|---|---|---|---|---|---|---|---|---|---|---|---|---|---|---|---|
| POST | 80 | 0 | 0 | 0 | 0 | 1 | 0 | 0 | 0 | 0 | 0 | 0 | 0 | 0 | 7 | 0 | 1 | 0 |
| NVB | 0 | 61 | 0 | 0 | 0 | 0 | 0 | 0 | 0 | 0 | 0 | 0 | 0 | 0 | 10 | 0 | 0 | 0 |
| VP | 0 | 0 | 97 | 1 | 0 | 0 | 0 | 0 | 0 | 0 | 0 | 0 | 0 | 3 | 5 | 0 | 0 | 0 |
| JJ | 0 | 0 | 2 | 66 | 0 | 0 | 0 | 0 | 0 | 0 | 0 | 0 | 0 | 0 | 7 | 0 | 1 | 0 |
| RB | 0 | 0 | 0 | 0 | 8 | 0 | 0 | 0 | 0 | 0 | 0 | 0 | 0 | 0 | 0 | 0 | 0 | 0 |
| RP | 2 | 0 | 0 | 0 | 0 | 43 | 2 | 0 | 0 | 0 | 0 | 0 | 0 | 0 | 0 | 0 | 0 | 0 |
| VNF | 0 | 0 | 0 | 0 | 0 | 3 | 57 | 0 | 0 | 0 | 0 | 0 | 0 | 0 | 4 | 0 | 0 | 0 |
| DET | 0 | 0 | 0 | 0 | 0 | 0 | 0 | 21 | 0 | 0 | 0 | 0 | 0 | 0 | 2 | 0 | 1 | 0 |
| VNN | 0 | 0 | 0 | 0 | 0 | 0 | 0 | 0 | 30 | 0 | 0 | 0 | 0 | 0 | 1 | 1 | 0 | 0 |
| PRP | 0 | 0 | 0 | 0 | 0 | 0 | 0 | 0 | 0 | 24 | 0 | 0 | 0 | 0 | 1 | 0 | 0 | 0 |
| NNF | 0 | 0 | 0 | 0 | 0 | 0 | 0 | 0 | 0 | 0 | 2 | 0 | 0 | 0 | 0 | 0 | 0 | 0 |
| JVB | 0 | 0 | 0 | 0 | 0 | 0 | 0 | 0 | 0 | 0 | 0 | 15 | 0 | 0 | 2 | 0 | 1 | 0 |
| NNPA | 0 | 0 | 0 | 0 | 0 | 0 | 0 | 0 | 0 | 0 | 0 | 0 | 9 | 0 | 0 | 0 | 1 | 0 |
| VFM | 0 | 0 | 2 | 0 | 0 | 0 | 0 | 0 | 0 | 0 | 0 | 0 | 0 | 30 | 1 | 0 | 0 | 0 |
| NNN | 5 | 1 | 2 | 2 | 0 | 2 | 0 | 0 | 0 | 0 | 0 | 2 | 0 | 0 | 377 | 0 | 5 | 0 |
| NNM | 0 | 0 | 0 | 0 | 0 | 0 | 0 | 0 | 0 | 0 | 0 | 0 | 0 | 0 | 2 | 27 | 0 | 0 |
| NNPI | 0 | 0 | 0 | 0 | 0 | 0 | 0 | 0 | 0 | 0 | 0 | 0 | 0 | 0 | 9 | 1 | 58 | 0 |
| CC | 0 | 0 | 0 | 0 | 0 | 0 | 0 | 0 | 0 | 0 | 0 | 0 | 0 | 0 | 0 | 0 | 0 | 18 |

Fig. IV Confusion matrix of the test results

## VIII. CONCLUSION AND FUTURE WORK

In this paper we have described two types of methods for dealing with unknown words in POS tagging Sinhala language. The application of the methods in Hidden Markov Model based part of speech tagging approach (which we have previously developed [1]) was evaluated and results were presented. The implementation was tested against 90551 words, 2754 sentences of Sinhala text corpus and that showed 91.5% accuracy in the tagging process with predicting tags to unknown words. So that the performance of the tagger prove that distinction between closed class and open class words and syntactic knowledge are reliable sources of information for handling unknown words in part of speech tagging of Sinhala language.

Though this research proposed a reliable approach to handle unknown words in POS tagging, further enhancements are required to improve and optimize the algorithm. Hence, several directions are suggested here for future work.

- Language specific morphological features can be helpful in predicting possible parts of speech for most of the verbs.

- Incorporation of named entity recognition techniques: information about identifying named entity could be possible clue in predicting parts of speech for most of the nouns.

- Instead of using only HMM, following a hybrid approach, with incurring above key knowledge to the system by a set of hand written rules.

- Increasing the size of the corpus and accuracy of tagged data.

ACKNOWLEDGMENT

We express our immense gratitude and many thanks to Dr. Harsha Kumara for his invaluable support in providing an initiative to this work. Many thanks to Ms. Kumudu Gamage at the Department of Linguistics, University of Kelaniya for her kind support.

**Apendixes 1:**

Sinhala Tag Set (Reference [8])

|    | Tag   | Description |
|----|-------|-------------|
| 1  | NNR   | Common Noun Root |
| 2  | NNM   | Common Noun Masculine |
| 3  | NNF   | Common Noun Feminine |
| 4  | NNN   | Common Noun Neuter |
| 5  | NNPA  | Proper Noun Animate |
| 6  | NNPI  | Proper Noun Inanimate |
| 7  | PRPM  | Pronoun Masculine |
| 8  | PRPF  | Pronoun Feminine |
| 9  | PRPN  | Pronoun Neuter |
| 10 | PRPC  | Pronoun Common |
| 11 | QFNUM | Number Quantifier |
| 12 | DET   | Determiner |
| 13 | JJ    | Adjective |
| 14 | RB    | Adverb |
| 15 | RP    | Particle |
| 16 | VFM   | Verb Finite Main |
| 17 | VNF   | Verb Non Finite |
| 18 | VP    | Verb Ptharticiple |
| 19 | VNN   | Verbal Non Finite Noun |
| 20 | POST  | Postpositions |
| 21 | CC    | Conjunctions |
| 22 | NVB   | Noun in Kriya Mula |
| 23 | JVB   | Adjective in Kriya Mula |
| 24 | UH    | Interjection |
| 25 | FRW   | Foreign Word |
| 26 | SYM   | Not Classified |